\documentclass[runningheads]{llncs}
\usepackage{graphicx}
\usepackage{amsmath,amssymb} % define this before the line numbering.
\usepackage{color}
\usepackage[width=122mm,left=12mm,paperwidth=146mm,height=193mm,top=12mm,paperheight=217mm]{geometry}
\usepackage{epsfig}
\usepackage{epstopdf}
\usepackage{multirow}
\usepackage[dvipsnames]{xcolor}

\usepackage[pagebackref=true,breaklinks=true,colorlinks,bookmarks=false]{hyperref}

\newcommand{\etal}{\emph{et al.}}

\usepackage{amsmath,amsfonts}
\usepackage{amssymb,amsopn}
\usepackage{bm} % bold symbol

\newcommand{\vct}[1]{\boldsymbol{#1}} % vector
\newcommand{\mat}[1]{\boldsymbol{#1}} % matrix
\newcommand{\cst}[1]{\mathsf{#1}}  % constant

\newcommand{\T}{^{\textrm T}} % transpose

\newcommand{\ProbOpr}[1]{\mathbb{#1}}
\newcommand{\var}[2]{%
\ifthenelse{\equal{#2}{}}{\ProbOpr{VAR}_{#1}}
{\ifthenelse{\equal{#1}{}}{\ProbOpr{VAR}\left[#2\right]}{\ProbOpr{VAR}_{#1}\left[#2\right]}}} % Expectation: syntax: V{1}{2} = V_1[2], V{}{2}=V[2], V{1}{} = V_1
\DeclareMathOperator{\argmax}{arg\,max}

\newcommand{\vtheta}{\vct{\theta}}

\newcommand{\vc}{\vct{c}}

\newcommand{\vx}{{\vct{x}}}

\newcommand{\vz}{{\vct{z}}}

\newcommand{\vo}{{\vct{o}}}

\newcommand{\vi}{\vct{i}}

\newcommand{\vf}{\vct{f}}
\newcommand{\vh}{\vct{h}}

\newcommand{\vphi}{\vct{\phi}}

\newcommand{\mW}{\mat{W}}

\newcommand{\mL}{\mat{L}}
\newcommand{\mI}{\mat{I}}

\newcommand{\cZ}{\cst{Z}}

\newcommand{\eat}[1]{}

\begin{document}

\pagestyle{headings}
\mainmatter

\titlerunning{Video Summarization with Long Short-term Memory}

\authorrunning{Ke Zhang, Wei-Lun Chao, Fei Sha, and Kristen  Grauman}

\title{Video Summarization with Long Short-term Memory} % Replace with your title

\author{Ke Zhang$^{1\star}$, Wei-Lun Chao$^{1}$\thanks{\hspace{4pt}Equal contributions}, Fei Sha$^{2}$, and Kristen  Grauman$^{3}$}

\institute{$^1$Dept. of Computer Science, U. of Southern California, United States\\
$^2$Dept. of Computer Science, U. of California, Los Angeles, United States\\
$^3$Dept. of Computer Science, U. of Texas at Austin, United States\\
\email{\{zhang.ke, weilunc\}@usc.edu, feisha@cs.ucla.edu, grauman@cs.utexas.edu}}

\maketitle

%%%%%%%%% ABSTRACT
\begin{abstract}
% !TEX root = main.tex

We propose a novel supervised learning technique for summarizing videos by  automatically selecting keyframes or key subshots. Casting the task as a structured prediction problem, our main idea is to use Long Short-Term Memory (LSTM) to model the variable-range temporal dependency among video frames, so as to derive both representative and compact video summaries. The proposed model successfully accounts for the sequential structure crucial to generating meaningful video summaries, leading to state-of-the-art results on two benchmark datasets.  
In addition to advances in modeling techniques, we introduce a strategy to address the need for a large amount of annotated data for training complex learning approaches to summarization.   There, our main idea is to exploit auxiliary annotated video summarization datasets, in spite of their heterogeneity in visual styles and contents.   Specifically, we show that domain adaptation techniques can improve learning by reducing the discrepancies in the original datasets' statistical properties.

\end{abstract}

%%%%%%%%% BODY TEXT
% !TEX root = main.tex
\section{Introduction}
\label{sIntro}

Video  has rapidly become one of the most common sources of visual information. The amount of video data is daunting --- it takes over 82 years to watch all videos uploaded to YouTube per day!  Automatic tools for analyzing and understanding video contents are thus essential. In particular, \textit{automatic video summarization} is a key tool to help human users browse video data.  A good video summary would compactly depict the original video, distilling its important events into a short watchable synopsis.  Video summarization can shorten video in several ways.  In this paper, we focus on the two most common ones: \emph{keyframe selection}, where the system identifies a series of defining frames~\cite{zhang1997integrated,gong14diverse,mundur2006keyframe,liu2010hierarchical,lee2012discovering} and \emph{key subshot selection}, where the system identifies a series of defining subshots, each of which is a temporally contiguous set of frames spanning a short time interval~\cite{ngo2003automatic,laganiere2008video,nam2002event,lu2013story}. 

There has been a steadily growing interest in studying learning techniques for video summarization. Many approaches are based on unsupervised learning, and define intuitive criteria to pick frames~\cite{zhang1997integrated,lee2012discovering,ngo2003automatic,lu2013story,hong2009event,khosla2013large,liu2002optimization,kang2006space,ma2002user} without explicitly  optimizing the evaluation metrics. Recent work has begun to explore supervised learning techniques~\cite{gong14diverse,Gygli2015video,zhang2016summary,gygli2014creating,chao2014large}. In contrast to unsupervised ones, supervised methods directly learn from human-created summaries to capture the underlying frame selection criterion as well as to output a subset of those frames that is more aligned with human semantic understanding of the video contents. 

Supervised learning for video summarization entails two questions: what type of learning model to use? and how to acquire enough annotated data for fitting those models? Abstractly, video summarization is a structured prediction problem:  the input to the summarization algorithm is a sequence of video frames, and the output is a binary vector indicating whether a frame is to be selected or not. This type of sequential prediction task is the underpinning of many popular algorithms for problems in speech recognition, language processing, etc. The most important aspect of this kind of task is that the decision to select cannot be made locally and in isolation --- the inter-dependency entails making decisions after considering all data from the original sequence. 

For video summarization, the inter-dependency across video frames is complex and highly inhomogeneous. This is not entirely surprising as human viewers rely on high-level semantic understanding of the video contents (and keep track of the unfolding of storylines) to decide whether a frame would be valuable to keep for a summary.  For example, in deciding what the keyframes are, \emph{temporally close} video frames are often visually similar and thus convey redundant information such that they should be condensed. However, the converse is not true.  That is, visually similar frames do not have to be temporally close. For example, consider summarizing the video ``leave home in the morning and come back to lunch at home and leave again and return to home at night." While the frames related to the ``at home'' scene can be visually similar, the semantic flow of the video dictates none of them should be eliminated.  Thus, a summarization algorithm that relies on examining visual cues only but fails to take into consideration the high-level semantic understanding about the video over a \emph{long-range temporal span} will erroneously eliminate important frames. 
Essentially, the nature of making those decisions is largely sequential -- any decision including or excluding frames is dependent on other decisions made on a temporal line.

Modeling variable-range dependencies where both short-range and long-range relationships intertwine is a long-standing  challenging problem in machine learning. Our work is inspired by the recent success of applying long short-term memory (LSTM) to structured prediction problems such as speech recognition~\cite{deng2013new,graves2013speech,graves2014towards} and image and video captioning~\cite{donahue2015long,yao2015describing,venugopalan2015sequence,venugopalan2015translating,karpathy2015deep}. LSTM is especially advantageous in modeling long-range structural dependencies where the influence by the distant past on the present and the future must be adjusted in a data-dependent manner. In the context of video summarization, LSTMs explicitly use its  memory cells to learn the progression of ``storylines'', thus to know when to forget or incorporate the past events to make decisions.   

In this paper, we investigate how to apply LSTM and its variants to supervised video summarization. We make the following contributions. We propose \textsf{vsLSTM}, a LSTM-based model for video summarization (Sec.~\ref{svsLSTM}).   Fig.~\ref{fig:LSTM_I} illustrates the conceptual design of the model.  We demonstrate that the sequential modeling aspect of LSTM is essential; the performance of multi-layer neural networks (MLPs) using neighboring frames as features is inferior.   We further show how LSTM's strength  can be enhanced by combining it with the determinantal point process (DPP), a recently introduced probabilistic model for diverse subset selection~\cite{gong14diverse,kulesza2012determinantal}. The resulting model achieves the best results on two recent challenging benchmark datasets (Sec.~\ref{sExp}). Besides advances in modeling, we also show how to address the practical challenge of insufficient human-annotated video summarization examples.  We show that model fitting can benefit from combining video datasets, despite their  heterogeneity in both contents and visual styles.  In particular, this benefit can be improved by ``domain adaptation'' techniques that aim to reduce the discrepancies in statistical characteristics across the diverse datasets.

The rest of the paper is organized as follows. Section~\ref{related} reviews related work of video summarization, and Section~\ref{sApproach} describes the proposed LSTM-based model and its variants. In Section~\ref{sExp}, we report empirical results.   We  examine our approach in several supervised learning settings and contrast it to other existing methods, and we analyze the impact of domain adapation for merging summarization datasets for training (Section~\ref{sec:da}).  We conclude our paper in Section~\ref{disc}.

% !TEX root = main.tex
\section{Related Work}
\label{related}

Techniques for automatic video summarization fall in two broad categories: unsupervised ones that rely on manually designed criteria to prioritize and select frames or subshots from videos~\cite{zhang1997integrated,mundur2006keyframe,lee2012discovering,ngo2003automatic,lu2013story,hong2009event,khosla2013large,liu2002optimization,ma2002user,de2011vsumm,furini2010stimo,li2010multi,potapov2014category,morere2015co,kim2014reconstructing,zhao2014quasi,Song2015TVSum,Chu2015video} and supervised ones that leverage human-edited summary examples (or frame importance ratings) to \emph{learn} how to summarize novel videos~\cite{gong14diverse,Gygli2015video,zhang2016summary,gygli2014creating,chao2014large}.  Recent results by the latter suggest great promise compared to traditional unupservised methods.

Informative criteria  include relevance~\cite{hong2009event,kang2006space,ma2002user,potapov2014category,Chu2015video}, representativeness or importance~\cite{lee2012discovering,ngo2003automatic,lu2013story,hong2009event,khosla2013large,kim2014reconstructing,Song2015TVSum}, and diversity or coverage~\cite{zhang1997integrated,liu2002optimization,de2011vsumm,li2010multi,zhao2014quasi}. Several recent methods also exploit auxiliary information such as web images~\cite{hong2009event,khosla2013large,kim2014reconstructing,Song2015TVSum} or video categories~\cite{potapov2014category} to facilitate the summarization process. 

Because they explicitly learn from human-created summaries, supervised methods are better equipped to align with how humans would summarize the input video.  For example, a prior supervised approach learns to combine multiple hand-crafted criteria so that the summaries are consistent with ground truth~\cite{Gygli2015video,gygli2014creating}.  Alternatively, the determinatal point process (DPP) --- a probabilistic model that characterizes how a representative and diverse subset can be sampled from a ground set --- is a valuable tool to model summarization in the supervised setting~\cite{gong14diverse,zhang2016summary,chao2014large}.

None of above work uses LSTMs to  model both the \emph{short-range and long-range dependencies} in the sequential video frames. The sequential DPP proposed in~\cite{gong14diverse} uses \emph{pre-defined temporal structures}, so the dependencies are ``hard-wired''. In contrast, LSTMs can model dependencies with a data-dependent on/off switch, which is extremely powerful for modeling sequential data~\cite{graves2013speech}.

LSTMs are used in~\cite{yang2015unsupervised} to model temporal dependencies to identify video highlights, cast as auto-encoder-based outlier detection. LSTMs are also used in modeling an observer's visual attention in analyzing images~\cite{xu2015show,jin2015aligning}, and to perform natural language video description~\cite{yao2015describing,venugopalan2015sequence,venugopalan2015translating}.   However, to the best of our knowledge, our work is the first to explore LSTMs for video summarization.  As our results will demonstrate, their flexibility in capturing sequential structure is quite promising for the task.

% !TEX root = main.tex
\section{Approach}
\label{sApproach}

In this section, we describe our methods for summarizing videos. We first formally state the problem and  the notations, and briefly review  LSTM \cite{gers2000learning,hochreiter1997long,zaremba2014learning}, the building block of our approach. We then introduce our first summarization model \textsf{vsLSTM}. Then we describe how we can enhance \textsf{vsLSTM} by combining it with a determinantal point process (DPP) that further takes the summarization structure (e.g., diversity among selected frames) into consideration. 

\subsection{Problem Statement}

We use $\vx= \{\vx_1, \vx_2, \cdots, \vx_t, \cdots, \vx_T\}$ to denote a sequence of  frames in a video to be summarized while $\vx_t$ is the visual features extracted at the $t$-th frame. 

The output of the summarization algorithm can take one of two forms.  The first is \emph{selected keyframes}~\cite{gong14diverse,mundur2006keyframe,liu2002optimization,de2011vsumm,furini2010stimo,wolf1996key}, where the summarization result is a subset of (isolated) frames.  The second is \emph{interval-based keyshots}~\cite{Gygli2015video,gygli2014creating,potapov2014category,Song2015TVSum}, where the  summary is a set of (short) intervals along the time axis.  Instead of binary information (being selected or not selected), certain datasets provide frame-level importance scores computed from human annotations~\cite{gygli2014creating,Song2015TVSum}. Those scores represent the likelihoods of the frames being selected as a part of summary.  Our models make use of all types of annotations --- binary keyframe labels, binary subshot labels, or frame-level importances --- as learning signals.\footnote{We describe below and in the Supplementary Material how to convert between the annotation formats when necessary.}  

Our models use frames as its internal representation. The inputs are frame-level features $\vx$ and the (target) outputs are either hard binary indicators or frame-level importance scores (i.e., softened indicators).

\subsection{Long Short-Term Memory (LSTM)}

LSTMs are a special kind of recurrent neural network that are adept at modeling long-range dependencies. 
At the core of the LSTMs are memory cells $\vc$ which encode, at every time step, the knowledge of the inputs that have been observed up to that step. The cells are modulated by nonlinear sigmoidal gates, and are applied multiplicatively. The gates determine whether the LSTM keeps the values at the gates (if the gates evaluate to $1$) or discard them (if the gates evaluate to $0$). 

\begin{figure}[t]
\centering
\begin{minipage}{.35\textwidth}
\centering
\includegraphics[width=\textwidth]{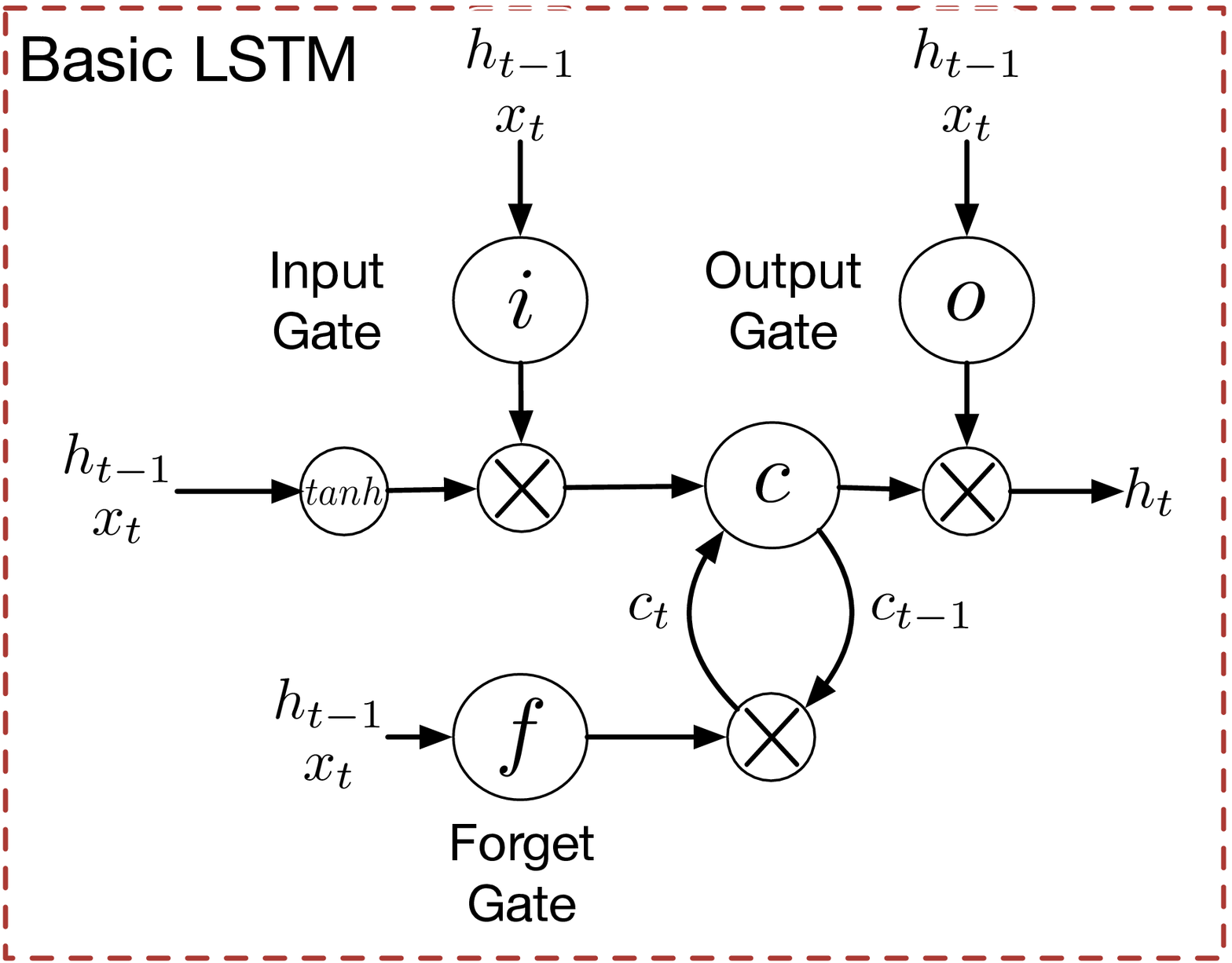}
\end{minipage}
\begin{minipage}{.5\textwidth}
\begin{align}
\label{LSTM}
\vi_t&=\textrm{sigmoid}(\mW_i[\vx_{t}\T,\vh_{t-1}\T]\T)\nonumber\\ 
\vf_t&=\textrm{sigmoid}(\mW_f[\vx_{t}\T,\vh_{t-1}\T]\T)\nonumber\\ 
\vo_t&=\textrm{sigmoid}(\mW_o[\vx_{t}\T,\vh_{t-1}\T]\T)\\ 
\vc_t&=\vi_t \odot \textrm{tanh}(\mW_c[\vx_{t}\T,\vh_{t-1}\T]\T)\nonumber \\
&\hspace{2em}  +\vf_t \odot \vc_{t-1}\nonumber\\
\vh_t&=\vo_t \odot \textrm{tanh}(\vc_t)\nonumber ,
\end{align}
\end{minipage}
\caption{\small The LSTM unit, redrawn from \cite{graves2014towards}. The memory cell is modulated jointly by the input, output and forget gates to control the knowledge transferred at each time step. $\odot$ denotes element-wise products.}
\label{fig:LSTM_Unit}
\end{figure}

There are three gates: the input gate $(\vi)$ controlling whether the LSTM considers its current input $(\vx_t)$, the forget gate $(\vf)$ allowing the LSTM to forget its previous memory $(\vc_{t})$, and the output gate $(\vo)$ deciding how much of the memory to transfer to the hidden states $(\vh_t)$. Together they enable the LSTM to learn complex long-term dependencies -- in particular, the forget date serves as a time-varying data-dependent on/off switch to selectively incorporating the past and present information. See Fig.~\ref{fig:LSTM_Unit} for a conceptual diagram of a LSTM unit and its algebraic definitions~\cite{graves2014towards}.  

\begin{figure}[t]
\centering
\includegraphics[height=1.5in]{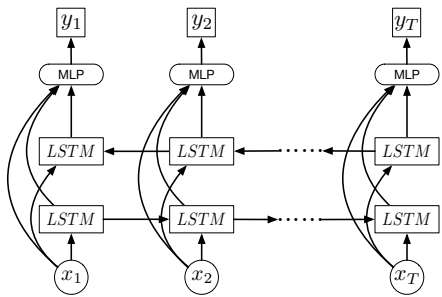}
\caption{\small Our \textsf{vsLSTM} model for video summarization. The model is composed of two LSTM (long short-term memory) layers: one layer models video sequences in the forward direction and the other the backward direction. Each \textit{LSTM} block is a LSTM unit, shown in Fig.~\ref{fig:LSTM_Unit}. The forward/backward chains model temporal inter-dependencies between the past and the future. The inputs to the layers are visual features extracted at frames. The outputs combine the LSTM layers' hidden states and the visual features with a multi-layer perceptron, representing the likelihoods of whether the frames should be included in the summary.  As our results will show, modeling sequential structures as well as the long-range dependencies is essential. }
\label{fig:LSTM_I}
\end{figure}

\subsection{{\textsf{vsLSTM}} for Video Summarization}
\label{svsLSTM}

Our \textsf{vsLSTM}  model is illustrated in Fig.~\ref{fig:LSTM_I}. There are several differences from the basic LSTM model. We use bidirectional LSTM layers~\cite{graves2005framewise} for modeling better long-range dependency in both the past and the future directions. Note that the forward and the backward chains do not directly interact.

We combine the information in those two chains, as well as the visual features,  with a multi-layer perceptron (MLP). The output of this perceptron is a scalar
\[
y_t = f_I(\vh_t^\text{forward}, \vh_t^\text{backward}, \vx_t).
\]

To learn the parameters in the LSTM layers and the MLP for $f_I(\cdot)$, our algorithm can use annotations in the forms of either the frame-level importance scores or the selected keyframes encoded as binary indicator vectors. In the former case, $y$ is a continuous variable and  in the latter case, $y$ is a binary variable.  The parameters are optimized with stochastic gradient descent.

\subsection{Enhancing {\textsf{vsLSTM}} by Modeling Pairwise Repulsiveness}
\label{sdppLSTM}

\begin{figure}[t]
\centering
\includegraphics[height=5cm]{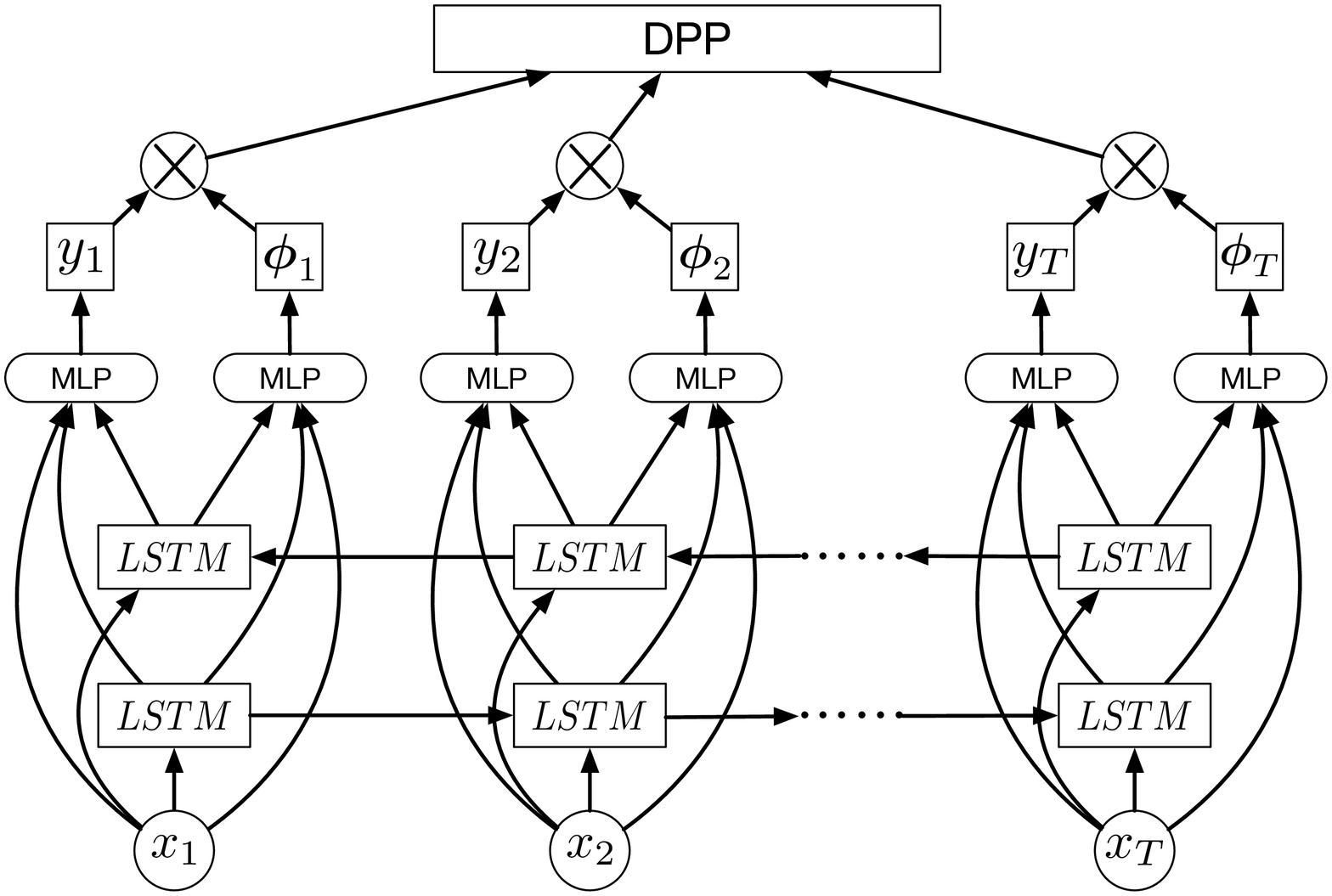}
\caption{\small Our \textsf{dppLSTM} model. It combines \textsf{vsLSTM} (Fig.~\ref{fig:LSTM_I}) and DPP by modeling both long-range dependencies and pairwise frame-level repulsiveness explicitly.}
\label{fig:LSTM_QD}
\end{figure}

\textsf{vsLSTM} excels at predicting the likelihood that a frame should be included or how important/relevant a frame is to the summary. We further enhance it with the ability to model pairwise frame-level ``repulsiveness''  by stacking it with a determinantal point process (DPP) (which we discuss in more detail below).   Modeling the repulsiveness aims to increase the diversity in the selected frames by eliminating redundant frames.  The modeling advantage provided in DPP has been exploited in DPP-based summarization methods~\cite{gong14diverse,zhang2016summary,chao2014large}. Note that diversity can only be measured ``collectively'' on a (sub)set of (selected) frames, not on frames independently or sequentially. The directed sequential nature in LSTMs is arguably \emph{weaker} in examining \emph{all the fames simultaneously} in the subset to measure diversity, thus is at the risk of having higher recall but lower precision. On the other hand, DPPs likely yield low recalls but high precisions. In essence, the two are complementary to each other.

\paragraph{\textbf{\textup{Determinantal point processes (DPP)}}} Given a ground set $\cZ$ of N items (e.g., all frames of a video), together with an N $\times$ N kernel matrix $\mL$ that records the pairwise frame-level similarity, a DPP encodes the probability to sample any subset from the ground set~\cite{gong14diverse,kulesza2012determinantal}.  The probability of a subset $\vz$ is proportional to the determinant of the corresponding principal minor of the matrix $\mL_{\vz}$
\begin{align}
P(\vz\subset\cZ; \mL) = \frac{\det(\mL_{\vz})}{\det(\mL+\mI)},
\end{align}
where $\mI$ is the N $\times$ N identity matrix. If two items are identical and appear in the subset,  $\mL_{\vz}$ will have identical rows and columns, leading to zero-valued determinant. Namely, we will have zero-probability assigned to this subset. A highly probable subset is one capturing significant diversity (i.e., pairwise dissimilarity). 

\paragraph{\textbf{\textsf{dppLSTM}}} Our \textsf{dppLSTM} model is schematically illustrated in Fig.~\ref{fig:LSTM_QD}. To exploit the strength of DPP in explicitly modeling diversity, we use the prediction of our \textsf{vsLSTM} in defining the $\mL$-matrix:
\begin{align}
L_{tt'} = y_t y_{t'} S_{tt'}= y_t y_{t'} \vphi_{t}\T \vphi_{t'},
\label{eDPPLDecompose}
\end{align}
where the similarity between the frames $x_t$ and $x_t'$ are modeled with the inner product of another multi-layer perceptron's outputs
\[
\vphi_t = f_S(\vh_t^\text{forward}, \vh_t^\text{backward}, \vx_t),\  \vphi_{t'} = f_S(\vh_{t'}^\text{forward}, \vh_{t'}^\text{backward}, \vx_{t'}). 
\]
This decomposition is similar in spirit to the  quality-diversity (QD) decomposition proposed in~\cite{kulesza2011learning}. While \cite{gong14diverse} also parameterizes $L_{tt'}$ with a single MLP, our model subsumes theirs. Moreover, our empirical results show that using two different sets of MLPs --- $f_I(\cdot)$ for frame-level importance and $f_S(\cdot)$ for similarity --- leads to better performance than using a single MLP to jointly model the two factors. (They  are implemented by one-hidden-layer neural networks with 256 sigmoid hidden units, and sigmoid and linear output units, respectively. See the Supplementary Material for details.)

\paragraph{\textbf{\textup{Learning}}} To train a complex model such as \textsf{dppLSTM}, we adopt a stage-wise optimization routine. We first train the MLP $f_I(\cdot)$ and the LSTM layers as in \textsf{vsLSTM}. Then, we train all the MLPs and the LSTM layers by maximizing the likelihood of keyframes specified by the DPP model.  Denote $\cZ^{(i)}$ as the collection of frames of the $i$-th video and $\vz^{(i)*}\subset\cZ^{(i)}$ as the corresponding target subset of keyframes.  We learn $\vtheta$ that parameterizes~(\ref{eDPPLDecompose}) by MLE~\cite{kulesza2012determinantal}:
\begin{equation}
\label{dpp_obj}
\vtheta^* = \argmax_{\vtheta} \sum_{i}\log\{P(\vz^{(i)*} \subset \cZ^{(i)}; \mL^{(i)}(\vtheta))\}.
\end{equation}
  Details are in the Supplementary Material. We have found this training procedure is effective in quickly converging to a good local optima.

\subsection{Generating Shot-based Summaries from Our Models}

Our \textsf{vsLSTM}  predicts frame-level importance scores, i.e., the likelihood that a frame should be included in the summary. For our \textsf{dppLSTM}, the approximate MAP inference algorithm~\cite{buchbinder2015tight} outputs a subset of selected frames.  Thus, for \textsf{dppLSTM} we use the procedure described in the Supplementary Material to convert them into keyshot-based summaries for evaluation. 

% !TEX root = main.tex
\section{Experimental Results}
\label{sExp}

We first define the experimental setting (datasets, features, metrics).  Then we provide key quantitative results demonstrating our method's advantages over existing techniques  (Sec.~\ref{sec:mainresult}).  Next we analyze more deeply the impact of our method design (Sec.~\ref{sec:analysis}) and explore the use of domain adaptation for ``homogenizing'' diverse summarization datasets (Sec.~\ref{sec:da}).  Finally, we present example qualitative results (Sec.~\ref{sec:qual}).

\subsection{Experimental Setup}\label{sec:setup}

\paragraph{\textbf{\textup{Datasets}}} We evaluate the performance of our models on two video datasets, \textbf{SumMe}~\cite{gygli2014creating} and \textbf{TVSum}~\cite{Song2015TVSum}.  \textbf{SumMe} consists of 25 user videos recording a variety of events such as holidays and sports. \textbf{TVSum} contains 50 videos downloaded from YouTube in 10 categories defined in the TRECVid Multimedia Event Detection (MED). Most of the videos are 1 to 5 minutes in length.

To combat the need of  a large amount of annotated data, we use two other annotated datasets whuch are annotated with keyframe-based summarization, \textbf{Youtube}~\cite{de2011vsumm} and \textbf{Open Video Project (OVP)}~\cite{openvideo,de2011vsumm}. We process them as\cite{gong14diverse} to create a ground-truth set of keyframes (then convert to a ground-truth sequence of frame-level importance scores) for each video.  We use the ground-truth in importance scores to train \textsf{vsLSTM} and convert the sequence to selected keyframes to train  \textsf{dppLSTM}. 

For evaluation, both datasets provide multiple user-annotated summaries for each video, either in the form of keyshots (\textbf{SumMe}) or frame-level importance scores (\textbf{TVSum}, convertible to keyshot-based summaries). Such conversions are documented in the Supplementary Material.

Table~\ref{tDataset} summarizes key characteristics of these datasets.  We can see that these four datasets are heterogeneous in both their visual styles and contents.

\begin{table*}[t]
\centering
\small
\caption{Key characteristics of datasets used in our empirical studies.}
\label{tDataset}
\begin{tabular}{c|c|c|c}\hline
\centering{Dataset} & \shortstack{\# of video} & \shortstack{Description} & \shortstack{Annotations}\\
\hline
\textbf{SumMe} & 25 & User generated videos of events & interval-based shots   \tabularnewline
\cline{1-3} 
\textbf{TVSum}  & 50 & YouTube videos (10 categories) & frame-level importance \tabularnewline
\hline
\textbf{OVP} & 50 & Documentary videos & selected keyframes  \tabularnewline
\cline{1-3} 
\textbf{YouTube} & 39 & YouTube videos (Sports, News, etc) & as summarization\tabularnewline
\hline
\end{tabular}
\end{table*}

\paragraph{\textbf{\textup{Features}}}  For most experiments, the feature descriptor of each frame is obtained by extracting the output of the penultimate  layer (pool 5) of the GoogLeNet model~\cite{szegedy2015going} (1024-dimensions). We also experiment with the same shallow features used in \cite{Song2015TVSum} (i.e., color histograms, GIST, HOG, dense SIFT) to provide a comparison to the deep features.

\paragraph{\textbf{\textup{Evaluation metrics}}}
Following the protocols in~\cite{Gygli2015video,gygli2014creating,Song2015TVSum}, we constrain the generated keyshot-based summary \textsf{A} to be less than 15\% in duration of the original video (details in the Supplementary Material). We then  compute the precision (P) and recall (R)  against  the user summary \textsf{B} for evaluation, according to the \textit{temporal overlap} between the two:
\begin{align}
& \text{P} =\frac{\text{\small overlapped duration of \textsf{A} and \textsf{B}}}{\text{\small duration of \textsf{A}}},
 \, \text{R} =\frac{\text{\small overlapped duration of \textsf{A} and \textsf{B}}}{\text{\small duration of \textsf{B}}},
\end{align}
as well as their harmonic mean F-score,
\begin{equation}
 \text{F} ={2\text{P}\times \text{R}}/{(\text{P}+\text{R})}\times 100\%.
\end{equation}
We also follow~\cite{Song2015TVSum,Gygli2015video} to compute the metrics when there are multiple human-annotated summaries of a video.

\paragraph{\textbf{\textup{Variants of supervised learning settings}}}
We study several settings for supervised learning, summarized in Table~\ref{tSetting}:
\begin{itemize}
\item {\textsf{Canonical}}\ This is the standard supervised learning setting where the training, validation, and testing sets are from the same dataset, though they are disjoint.
\item {\textsf{Augmented}}\  In this setting, for a given dataset,  we randomly leave 20\% of it for testing, and augment the remaining 80\%  with the other three datasets to form an augmented training and validation dataset. Our hypothesis is that, despite being heterogeneous in styles and contents, the augmented dataset can be beneficial in improving the performance of our models because of the increased amount of annotations.  
\item {\textsf{Transfer}}\  In this setting, for a given dataset, we use the other three datasets for training and validation and test the learned models on the dataset. We are interested in investigating if existing datasets can effectively transfer summarization models to new unannotated datasets. If the transfer can be successful, then it would be possible to summarize a large number of videos in the wild where there is virtually no closely corresponding annotation.
\end{itemize}

\begin{table}[t]
\centering
\small
\caption{Supervision settings tested }
\label{tSetting}
\begin{tabular}{c|c|c|c}\hline
Dataset & Settings & Training \& Validation & Testing\\
\hline 
\multirow{3}{*}{SumMe} & \textsf{Canonical} & $80\%$ SumMe & $20\%$ SumMe \\ \cline{2-4} 
&  \textsf{Augmented} & OVP + Youtube + TVSum + $80\%$ SumMe & $20\%$ SumMe \\ \cline{2-4}
&  \textsf{Transfer} & OVP + Youtube + TVSum & SumMe \\ \cline{2-4}
\hline
\multirow{3}{*}{TVSum} &  \textsf{Canonical}  & $80\%$ TVSum & $20\%$ TVSum \\ \cline{2-4}
&  \textsf{Augmented}  & OVP + Youtube + SumMe + $80\%$ TVSum & $20\%$ TVSum \\ \cline{2-4}

&  \textsf{Transfer} & OVP + Youtube + SumMe & TVSum \\ \cline{2-4}
\hline
\end{tabular}
\end{table} 

\subsection{Main Results}\label{sec:mainresult}

Table~\ref{tMain} summarizes the performance of our methods and contrasts to those attained by prior work. Red-colored numbers indicate that our dppLSTM obtains the best performance in the corresponding setting. Otherwise the best performance is bolded. In the common setting of ``\textsf{Canonical}'' supervised learning, on TVSum, both of our two methods outperform the state-of-the-art. However, on SumMe, our methods underperform the state-of-the-art, likely due to the fewer annotated training samples in SumMe.

What is particularly interesting is that our methods can be significantly improved when the amount of annotated data is increased. In particular, in the case of \textsf{Transfer} learning, even though the three training datasets are significantly different from the testing dataset, our methods leverage the annotations effectively to improve accuracy over the \textsf{Canonical} setting, where the amount of annotated training data is limited. The best performing setting is \textsf{Augmented}, where we combine all four datasets together to form one training dataset. 

The results suggest that with sufficient annotated data, our model can capture temporal structures better than prior methods that lack explicit temporal structures~\cite{khosla2013large,Gygli2015video,gygli2014creating,li2010multi,Song2015TVSum} as well as those that consider only pre-defined ones~\cite{gong14diverse,zhang2016summary}. More specifically, bidirectional LSTMs and DPPs help to obtain diverse results conditioned on the whole video while leveraging the sequential nature of videos.  See the Supplementary Material for further discussions.

\begin{table}[t]
\centering
\small
\caption{Performance (F-score) of various video summarization methods. Published results are denoted in~\textbf{\textit{bold italic}}; our implementation is in normal font.  Empty boxes indicate settings inapplicable to the method tested.}
\label{tMain}
\begin{tabular}{c|c|c|c|c|c}
\hline 
Dataset & Method & \textsf{unsupervised} & \textsf{Canonical} & \textsf{Augmented} &\textsf{Transfer} \\
\hline 
\multirow{6}{*}{SumMe} & \cite{li2010multi}  & \textbf{\textit{26.6}} & & & \\ \cline{2-6}
& \cite{gygli2014creating} & & \textbf{\textit{39.4}}  & & \\ \cline{2-6}
& \cite{Gygli2015video} & & \textbf{\textit{39.7}} & & \\ \cline{2-6}
& \cite{zhang2016summary} & & \textbf{\textit{40.9}}$^\dagger$ & 41.3 & 38.5 \\ \cline{2-6}
& \textsf{vsLSTM (ours)} & & 37.6\scriptsize{$\pm$0.8}& 41.6\scriptsize{$\pm$0.5}& 40.7\scriptsize{$\pm$0.6}  \\ \cline{2-6}
& \textsf{dppLSTM (ours)} & & 38.6\scriptsize{$\pm$0.8}& \color{red}{{42.9}}\scriptsize{$\pm$0.5}& {\color{red}41.8\scriptsize{$\pm$0.5}}   \\ \cline{2-6}
 \hline
\hline
\multirow{5}{*}{TVSum} &  \cite{zhao2014quasi}&  \textbf{\textit{46.0}} & & & \\ \cline{2-6}
&  \cite{khosla2013large}$^\ddagger$ & \textbf{\textit{36.0}} & & &  \\ \cline{2-6}
&  \cite{Song2015TVSum}$^\ddagger$ & \textbf{\textit{50.0}} & & &  \\ \cline{2-6}
& \textsf{vsLSTM (ours)} & & 54.2\scriptsize{$\pm$0.7}& 57.9\scriptsize{$\pm$0.5}& 56.9\scriptsize{$\pm$0.5}  \\ \cline{2-6}
& \textsf{dppLSTM (ours)}  & & {\color{red}54.7\scriptsize{$\pm$0.7}}& \color{red}{{59.6}}\scriptsize{$\pm$0.4}& {\color{red}58.7\scriptsize{$\pm$0.4}}  \\ \cline{2-6}
 \cline{2-6}
\hline
\end{tabular}
\begin{flushleft}
{\scriptsize $^\dagger$: build video classifiers using TVSum~\cite{Song2015TVSum}. $^\ddagger$: use auxiliary web images for learning.}
\end{flushleft}
\end{table}

\subsection{Analysis}\label{sec:analysis}

\begin{table}[t]
\centering
\small
\caption{Modeling video data with LSTMs is beneficial. The reported numbers are F-scores by various summarization methods.}
\label{tSeq}
\begin{tabular}{c|l|c|c|c}
\hline 
Dataset & Method & \textsf{Canonical}& \textsf{Augmented} & \textsf{Transfer} \\
\hline 
\multirow{3}{*}{SumMe} 
& \textsf{MLP-Shot} & {\textbf{39.8}}\scriptsize{$\pm$0.7} & 40.7\scriptsize{$\pm$0.7} & 39.8\scriptsize{$\pm$0.6}  \\ \cline{2-5}
& \textsf{MLP-Frame} & 38.2\scriptsize{$\pm$0.8} & 41.2\scriptsize{$\pm$0.8}& 40.2\scriptsize{$\pm$0.9}  \\ \cline{2-5}
& \textsf{vsLSTM} & 37.6\scriptsize{$\pm$0.8} & {\color{red}41.6\scriptsize{$\pm$0.5}}& {\color{red}40.7\scriptsize{$\pm$0.6}} \\ \cline{2-5}
\hline \hline
\multirow{3}{*}{TVSum} 
& \textsf{MLP-Shot} & {\textbf{55.2}}\scriptsize{$\pm$0.5} & 56.7\scriptsize{$\pm$0.5}& 55.5\scriptsize{$\pm$0.5}  \\ \cline{2-5}
& \textsf{MLP-Frame} & 53.7\scriptsize{$\pm$0.7}& 56.1\scriptsize{$\pm$0.7} & 55.3\scriptsize{$\pm$0.6}   \\ \cline{2-5}
& \textsf{vsLSTM} & 54.2\scriptsize{$\pm$0.7}& {\color{red}57.9\scriptsize{$\pm$0.5}}& {\color{red}56.9\scriptsize{$\pm$0.5}}  \\ \cline{2-5}
\hline
\end{tabular}
\end{table}

Next we analyze more closely several settings of interest.

\paragraph{\textbf{\textup{How important is sequence modeling?}}} Table~\ref{tSeq} contrasts the performance of the LSTM-based method \textsf{vsLSTM} to a multi-layer perceptron based baseline. In this baseline,  we learn a two-hidden-layer MLP that has the same number of hidden units in each layer as does one of the MLPs of our model. 

Since MLP cannot explicitly capture temporal information, we consider two variants in the interest of fair comparison to our LSTM-based approach.  In the first variant \textsf{MLP-Shot}, we use the averaged frame features in a shot as the inputs to the MLP and predict shot-level importance scores. The ground-truth shot-level importance scores are derived as the average of the corresponding frame-level importance scores. The predicted shot-level importance scores are then used to select keyshots and the resulting shot-based summaries are then compared to user annotations.  In the second variant \textsf{MLP-Frame}, we concatenate all visual features within a $K$-frame ($K = 5$ in our experiments) window centered around each frame to be the inputs for predicting frame-level importance scores. 

It is interesting to note that in the \textsf{Canonical} setting, MLP-based approaches outperform \textsf{vsLSTM}. However, in all other settings where the amount of annotations is increased, our \textsf{vsLSTM} is able to outperform the MLP-based methods noticeably. This confirms the common perception about LSTMs: while they are powerful, they often demand a larger amount of annotated data in order to perform well.

\paragraph{\textbf{\textup{Shallow versus deep features?}}}

We also study the effect of using alternative visual features for each frame. Table~\ref{tShallow} suggests that deep features are able to modestly improve performance over the shallow features. Note that our~\textsf{dppLSTM} with shallow features still outperforms~\cite{Song2015TVSum}, which reported results on TVSum using the same shallow features (i.e., color histograms, GIST, HOG, dense SIFT). 

\begin{table}[t]
\centering
\small
\caption{Summarization results (in F-score) by our \textsf{dppLSTM} using shallow and deep features. Note that~\cite{Song2015TVSum} reported~\textbf{\textit{50.0\%}} on TVSum using the same shallow features.}
\label{tShallow}
\begin{tabular}{c|c|c|c}
\hline 
Dataset & Feature type & \textsf{Canonical}& \textsf{Transfer}\\
\hline 
\multirow{2}{*}{SumMe} & deep  & 38.6\scriptsize{$\pm$0.8}& {\textbf{41.8}}\scriptsize{$\pm$0.5}  \\ \cline{2-4}
& shallow  & 38.1\scriptsize{$\pm$0.9}& 40.7\scriptsize{$\pm$0.5} \\ \cline{2-4}
\hline \hline
\multirow{2}{*}{TVSum} & deep & 54.7\scriptsize{$\pm$0.7} &  {\textbf{58.7}}\scriptsize{$\pm$0.4}  \\ \cline{2-4}
& shallow & 54.0\scriptsize{$\pm$0.7}& 57.9\scriptsize{$\pm$0.5}  \\ \cline{2-4}
\hline
\end{tabular}
\end{table}

\begin{table}[t]
\centering
\small
\caption{\small Results by \textsf{vsLSTM} on different types of annotations in the \textsf{Canonical} setting}
\label{tAnnotation}
\begin{tabular}{c|c|c}\hline
dataset & Binary & Importance score\\ \hline
SumMe & 37.2\scriptsize{$\pm$0.8} & 37.6\scriptsize{$\pm$0.8}\\ \hline
TVSum & 53.7\scriptsize{$\pm$0.8} & 54.2\scriptsize{$\pm$0.7}\\ \hline
\end{tabular}
\end{table}

\paragraph{\textbf{\textup{What type of annotation is more effective?}}} There are two common types of annotations in video summarization datasets: binary indicators of whether a frame is selected or not and frame-level importance scores on how likely a frame should be included in the summary. While our models can take either format, we suspect the frame-level importance scores  provide richer information than the binary indicators as they represent relative goodness among frames.. 

Table~\ref{tAnnotation} illustrates the performance of our \textsf{vsLSTM} model when using the two different annotations, in the \textsf{Canonical} setting.  Using frame-level importance scores has a consistent advantage. 

However, this does not mean binary annotation/keyframes annotations cannot be exploited. Our \textsf{dppLSTM} exploits both frame-level importance scores and binary signals. In particular, \textsf{dppLSTM} first uses frame-level importance scores to train its LSTM layers and then uses binary indicators to form objective functions to fine tune (cf. Section~\ref{sApproach} for the details of this stage-wise training). Consequently, comparing the results in Table~\ref{tMain} to Table~\ref{tAnnotation}, we see that \textsf{dppLSTM} improves further by utilizing both types of annotations.

\subsection{Augmenting the Training Data with Domain Adaptation}\label{sec:da}

While Table~\ref{tMain} clearly indicates the advantage of augmenting the training dataset, those auxiliary datasets are often different from the target one in contents and styles. We improve summarization further by borrowing the ideas from visual domain adaptation for object recognition~\cite{saenko2010adapting,gong2012geodesic,donahue2013decaf}. The main idea is  first eliminate the discrepancies in data distribution  before augmenting.

Table~\ref{tAdaptTran} shows the effectiveness of this idea. We use a simple domain adaptation technique~\cite{sun2016return} to  reduce the data distribution discrepancy among all four datasets, by transforming the visual features linearly such that the covariance matrices for the four datasets are close to each other.  The ``homogenized'' datasets, when combined (in both the \textsf{Transfer} and \textsf{Augmented} settings), lead to an improved summary F-score. The improvements are especially pronounced for the smaller dataset SumMe.

\begin{table}[t]
\centering
\small
\caption{Summarization results by our  model in the \textsf{Transfer} and \textsf{Augmented} settings, optionally with visual features linearly adapted  to reduce cross-dataset discrepancies}
\label{tAdaptTran}
\begin{tabular}{c|l|c|c||c|c}
\hline 
\multirow{2}{*}{Dataset} & \multirow{2}{*}{Method} & \multicolumn{2}{|c||}{\textsf{Transfer}} & \multicolumn{2}{|c}{\textsf{Augmented}} \\ \cline{3-6}
& & {w/o Adaptation}  & {w/ Adaptation} & {w/o Adaptation}  & {w/ Adaptation} \\
\hline 
\multirow{2}{*}{SumMe} 
& \textsf{vsLSTM} & 40.7\scriptsize{$\pm$0.6} & 41.3\scriptsize{$\pm$0.6} & 41.6\scriptsize{$\pm$0.5} & 42.1\scriptsize{$\pm$0.6} \\ \cline{2-6}
& \textsf{dppLSTM}  & 41.8\scriptsize{$\pm$0.5} & 43.1\scriptsize{$\pm$0.6} & 42.9\scriptsize{$\pm$0.5} & 44.7\scriptsize{$\pm$0.5} \\
\hline
\multirow{2}{*}{TVSum} 
& \textsf{vsLSTM} &  56.9\scriptsize{$\pm$0.5} &  57.0\scriptsize{$\pm$0.5} &  57.9\scriptsize{$\pm$0.5} &  58.0\scriptsize{$\pm$0.5} \\ \cline{2-6}
& \textsf{dppLSTM} & 58.7\scriptsize{$\pm$0.4} & 58.9\scriptsize{$\pm$0.4} & 59.6\scriptsize{$\pm$0.4} & 59.7\scriptsize{$\pm$0.5}\\
\hline
\end{tabular}
\end{table}

% !TEX root = exp_result.tex
\subsection{Qualitative Results}\label{sec:qual}

\begin{figure}[t]
\centering
\includegraphics[width=8.6cm]{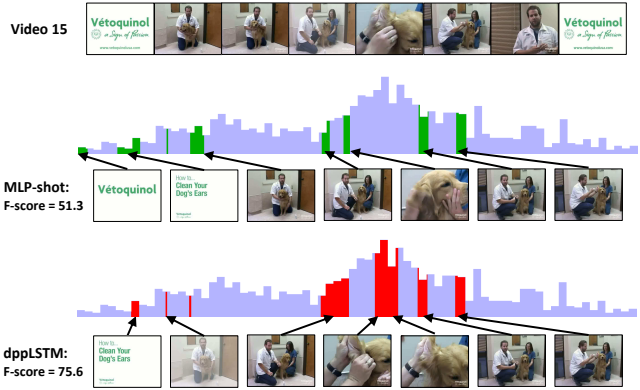}
\caption{\small Exemplar video summaries by \textsf{MLP-Shot} and \textsf{dppLSTM}, along with the ground-truth importance ({\color{blue}blue} background). See texts for details. We index videos as in \cite{Song2015TVSum}.}
\label{fig:qual1}
\end{figure}

We provide exemplar video summaries in Fig.~\ref{fig:qual1}. We illustrate the temporal modeling capability of \textsf{dppLSTM} and contrast with \textsf{MLP-Shot}. 

The height of the {\color{blue}blue} background indicates the ground-truth frame-level importance scores of the video. The marked {\color{red}red} and {\color{green}green} intervals are the ones selected by~\textsf{dppLSTM} and \textsf{MLP-Shot} as the summaries, respectively. \textsf{dppLSTM} can capture temporal dependencies and thus identify the most important part in the video, i.e. the frame depicting  the cleaning of the dog's ears. \textsf{MLP-Shot}, however, completely misses  selecting such subshots even though those subshots have much higher ground-truth importance scores than the neighboring frames.  We believe this is because \textsf{MLP-Shot} does not capture the sequential semantic flow properly and lacks the knowledge that if the neighbor frames are important, then the frames in the middle could be important too.

It is also very interesting to note that despite the fact that DPP models usually eliminate similar elements, \textsf{dppLSTM} can still select similar but important subshots: subshots of two people with dogs before and after  cleaning the dog's ear are both selected. This highlights \textsf{dppLSTM}'s ability to adaptively model long-range (distant states) dependencies. 

\begin{figure}[t]
\centering
\includegraphics[width=8.6cm]{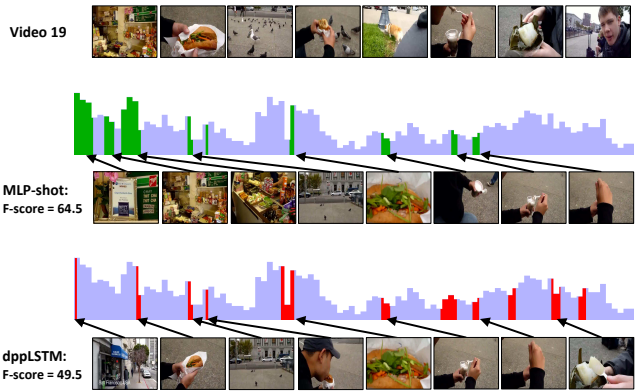}
\caption{\small A failure case by \textsf{dppLSTM}. See texts for details. We index videos as in \cite{Song2015TVSum}.}
\label{fig:qual2}
\end{figure}

Fig.~\ref{fig:qual2} shows a failure case of \textsf{dppLSTM}. This video is an outdoor ego-centric video and  records very diverse contents. In particular, the scenes change among a sandwich shop, building, food, and the town square. From the summarization results we see that \textsf{dppLSTM} still selects diverse contents, but fails to capture the beginning frames --- those frames all have high importance scores and are visually similar  but are \emph{temporally clustered crowdedly}.  In this case, \textsf{dppLSTM} is forced to eliminate some of them, resulting in low recall.  On the other hand, \textsf{MLP-Shot} needs only to predict importance scores without being diverse, which leads to higher recall and F-scores. Interestingly, \textsf{MLP-Shot} predicts poorly towards the end of the video, whereas the repulsiveness modeled by \textsf{dppLSTM} gives the method an edge to select a few frames in the end of the video.

In summary, we expect our approaches to work well on videos whose contents change smoothly (at least within a short interval) such that the temporal structures can be well captured. For videos with rapid changing and  diverse contents,  higher-level semantic cues (e.g., object detection as in~\cite{lee2012discovering,lu2013story}) could be complementary and should be incorporated.
% !TEX root = main.tex
\section{Conclusion}
\label{disc}

Our work explores Long Short-Term Memory to develop novel supervised learning approaches to automatic video summarization. Our LSTM-based models outperform competing methods on two challenging benchmarks. There are several key contributing factors: the modeling capacity by LSTMs to capture variable-range inter-dependencies, as well as our idea to complement LSTMs' strength with DPP to explicitly model inter-frame repulsiveness to encourage diverse selected frames. While LSTMs  require a large number of annotated samples, we show how to mediate this demand by exploiting the existence of other annotated video datasets, despite their heterogeneity in style and content.  Preliminary results are very promising, suggesting future research directions of developing more sophisticated techniques that can bring together a vast number of available video datasets for video summarization.  In particular, it would be very productive to explore new sequential models that can enhance LSTMs' capacity in modeling video data, by learning to encode semantic understanding of video contents and using them to guide summarization and other tasks in visual analytics.   
\\[1em]
\textbf{Acknowledgements} {\small KG is partially supported by NSF IIS-1514118 and a gift from Intel. Others are partially supported by USC Graduate Fellowships, NSF IIS-1451412, 1513966, CCF-1139148 and A. P. Sloan Research Fellowship.}

% \clearpage
\bibliographystyle{splncs}
\bibliography{main}

\appendix
\title{Supplementary Material: Video Summarization with Long Short-term Memory}
\author{}
\institute{}
\maketitle
In this Supplementary Material, we provide details omitted in the main text:
\begin{itemize}
\item Section~\ref{sConvert}: converting between different formats of ground-truth annotations (Section~3.1 in the main text)
\item Section~\ref{sData}: details of the datasets (Section~4.1 in the main text) 
\item Section~\ref{sLSTM}: details of our LSTM-based models, including the learning objective for~\textsf{dppLSTM} and the generating process of shot-based summaries for both~\textsf{vsLSTM} and~\textsf{dppLSTM} (Section~3.4 and 3.5 in the main text)
\item Section~\ref{sStruct}: comparing different network structures for~\textsf{dppLSTM} (Section~3.4 in the main text)
\item Section~\ref{sOther}: Other implementation details
\item Section~\ref{Disc_VS}: Additional discussions on video summarization
\end{itemize}

\section{Converting between different formats of ground-truth annotations}
\label{sConvert}
As mentioned in Section~3.1 of the main text, existing video summarization datasets usually provide the ground-truth annotations in (one of) the following three formats --- (a) selected keyframes, (b) interval-based keyshots, and (c) frame-level importance scores. See Table~\ref{annotations} for illustration.

In order to combine multiple datasets to enlarge the training set, or to enable any (supervised) video summarization algorithm to be trained under different ground-truth formats, we introduce a general procedure to convert between different formats. \emph{Note that we perform this procedure to the ground truths only in the training phase.} In the testing phase, we directly compare with the user-generated summaries in their original formats, unless stated otherwise (see Section~\ref{sData}). Also note that certain conversions require \emph{temporal segmentation} to cut a video into disjoint time intervals, where each interval contains frames of similar contents. Since none of the datasets involved in the experiments provides \emph{ground-truth temporal segmentation}, we apply the kernel temporal segmentation (KTS) proposed by Potapov~\etal~\cite{potapov2014category}. The resulting intervals are around 5 seconds on average.

\begin{table*}
\centering
\small
\caption{Illustration of different formats of ground-truth annotations for video summarization. We take a 6-frame sequence as an example.}
\label{annotations}
\vskip 0.25em
\begin{tabular}{l|l}
\centering{Format} & Description \\
\hline
(a) keyframes & \{frame 2, frame 6\} or [0 1 0 0 0 1]\\
\hline
(b) interval-based keyshots & \{frames 1--2, frames 5--6\} or [1 1 0 0 1 1]\\
\hline
(c) frame-level importance scores & [0.5 0.9 0.1 0.2 0.7 0.8]\\
\hline
\end{tabular}
\vskip -0.5em
\end{table*} 

\subsection{keyframes $\rightarrow$ keyshots and frame-level scores}
\label{keyframe}
To covert keyframes into keyshots, we first \emph{temporally segment} a video into disjoint intervals using KTS~\cite{potapov2014category}. Then if an interval contains at least one keyframe, we view such an interval as a keyshot, and mark all frames of it with score 1; otherwise, 0.

To prevent generating too many keyshots, we rank the candidate intervals (those with at least one keyframe) in the descending order by the number of key frames each interval contains divided by its duration. We then select intervals in order so that the total duration of keyshots is below a certain threshold (e.g., using the knapsack algorithm as in~\cite{Song2015TVSum}).

\subsection{keyshots $\rightarrow$ keyframes and frame-level scores}
\label{keyshot}
Given the selected keyshots, we can randomly pick a frame, or pick the middle frame, of each keyshot to be a keyframe. We also directly mark frames contained in keyshots with score 1. For those frames not covered by any keyshot, we set the corresponding importance scores to be 0.

\subsection{frame-level scores $\rightarrow$ keyframes and keyshots}
\label{score}
To convert frame-level importance scores into keyshots, we first perform~\emph{temporal segmentation}, as in Section~\ref{keyframe}. We then compute interval-level scores by averaging the scores of frames within each interval. We then rank intervals in the descending order by their scores, and select them in order so that the total duration of keyshots is below a certain threshold (e.g., using the knapsack algorithm as in~\cite{Song2015TVSum}). We further pick the frame with the highest importance score within each keyshot to be a keyframe.
\newline
\newline
Table~\ref{conversion} summarizes the conversions described above.

\begin{table*}
\centering
\small
\caption{Illustration of the converting procedure described in Section~\ref{keyframe}--\ref{score}. We take a 6-frame sequence as an example, and assume that the temporal segmentation gives three intervals, \{frames 1--2, frames 3--4, frames 5--6\}. The threshold of duration is 5.}
\label{conversion}
\vskip 0.25em
\begin{tabular}{l|l}
Conversion & \multicolumn{1}{|c}{Description} \\
\hline
Section~\ref{keyframe} & (a) [0 1 0 0 0 1] $\rightarrow$ (b) [1 1 0 0 1 1], (c) [1 1 0 0 1 1] \\
\hline
Section~\ref{keyshot} & (b) [1 1 0 0 1 1] $\rightarrow$ (a) [0 1 0 0 0 1], (c) [1 1 0 0 1 1] \\
\hline
Section~\ref{score} &  (c) [0.5 0.9 0.1 0.2 0.7 0.8] $\rightarrow$ (b) [1 1 0 0 1 1], (a) [0 1 0 0 0 1]\\
\hline
\end{tabular}
\vskip -0.5em
\begin{flushleft}
\vskip -.5em
\centering
(a) keyframes (b) interval-based keyshots (c) frame-level importance scores
\end{flushleft}
\end{table*}

\section{Details of the datasets}
\label{sData}

In this section, we provide more details about the four datasets --- \textbf{SumMe}~\cite{gygli2014creating}, \textbf{TVSum}~\cite{Song2015TVSum}, \textbf{OVP}~\cite{openvideo,de2011vsumm}, and \textbf{Youtube}~\cite{de2011vsumm} --- involved in the experiments. Note that \textbf{OVP} and \textbf{Youtube} are only used to augment the training set.

\subsection{Training ground truths}
Table~\ref{tDataset} lists the training and testing ground truths provided in each dataset. Note that in training, we require a single ground truth for each video, which is directly given in \textbf{SumMe} and \textbf{TVSum}, but not in \textbf{OVP} and \textbf{Youtube}. We thus follow~\cite{gong14diverse} to create a single ground-truth set of keyframes from multiple user-annotated ones for each video.

Table~\ref{tAlgorithm} summarizes the formats of training ground truths required by our proposed methods (\textsf{vsLSTM}, \textsf{dppLSTM}) and baselines (\textsf{MLP-Shot}, \textsf{MLP-Frame}). We perform the converting procedure described in Section~\ref{sConvert} to obtain the required training formats if they are not provided in the dataset. We perform KTS~\cite{potapov2014category} for temporal segmentation for all datasets.

\subsection{Testing ground truths for TVSum}
\textbf{TVSum} provides for each video multiple sequence of frame-level importance scores annotated by different users. We follow~\cite{Song2015TVSum} to convert each sequence into a keyshot-based summary for evaluation, which is exactly the one in Section~\ref{score}. We set the threshold to be 15\% of the original video length, following~\cite{Song2015TVSum}.

\begin{table*}
\centering
\small
\caption{Training and testing ground truths provided for each video in the datasets.}
\label{tDataset}
\vskip 0.25em
\begin{tabular}{c|c|c}
\multicolumn{1}{c|}{Dataset} & Training ground truths & Testing ground truths\\
\hline
SumMe & a sequence of frame-level scores & multiple sets of keyshots \tabularnewline
\hline
TVSum &  a sequence of frame-level scores & multiple sequences of frame-level scores$^\dagger$ \tabularnewline
\hline
OVP & multiple sets of keyframes$^\ddagger$  & - \tabularnewline
\hline
Youtube &  multiple sets of keyframes$^\ddagger$ & - \tabularnewline
\hline
\end{tabular}
\vskip -0.5em
\begin{flushleft}
\vskip -.5em
$^\dagger$ \small following~\cite{Song2015TVSum}, we convert the frame-level scores into keyshots for evaluation. \\
$^\ddagger$ \small following~\cite{gong14diverse}, we create a single ground-truth set of keyframes for each video.
\end{flushleft}
\end{table*} 

\begin{table*}
\centering
\small
\caption{The formats of training ground truths required by~\textsf{vsLSTM}, \textsf{dppLSTM}, \textsf{MLP-Shot}, and \textsf{MLP-Frame}.}
\label{tAlgorithm}
\vskip 0.25em
\begin{tabular}{c|c}
\multicolumn{1}{c|}{Method} & Training ground truths \\
\hline
\textsf{MLP-Shot} & shot-level importance scores$^\dagger$  \tabularnewline
\hline
\textsf{MLP-Frame} & frame-level importance scores\tabularnewline
\hline
\textsf{vsLSTM} & frame-level importance scores \tabularnewline
\hline
\textsf{dppLSTM} &  keyframes, frame-level importance scores$^\ddagger$ \tabularnewline
\hline
\end{tabular}
\vskip -0.5em
\begin{flushleft}
\vskip -.5em
$^\dagger$ The shot-level importance scores are derived as the averages of the corresponding frame-level importance scores. We perform KTS~\cite{potapov2014category} to segment a video into shots (disjoint intervals). \\
$^\ddagger$ We pre-train the MLP $f_I(\cdot)$ and the LSTM layers using frame-level scores.
\end{flushleft}
\end{table*} 

\section{Details of our LSTM-based models}
\label{sLSTM}
In this section, we provide more details about the proposed LSTM-based models for video summarization.

\subsection{The learning objective of dppLSTM}
As mentioned in Section~3.4 of the main text, we adopt a stage-wise optimization routine to learn \textsf{dppLSTM} --- the first stage is based on the prediction error of importance scores; the second stage is based on the maximum likelihood estimation (MLE) specified by DPPs. Denote $\cZ$ as a ground set of N items (e.g, all frames of a video), and $\vz^*\subset\cZ$ as the target subset (e.g., the subset of keyframes). Given the N $\times$ N kernel matrix $\mL$, the probability to sample $\vz^*$ is
\begin{equation}
P(\vz^*\subset\cZ; \mL) = \frac{\det(\mL_{\vz^*})}{\det(\mL+\mI)},
\end{equation}
where $\mL_{\vz^*}$ is the principal minor indexed by $\vz^*$, and $\mI$ is the N $\times$ N identity matrix.

In \textsf{dppLSTM}, $\mL$ is parameterized by $\vtheta$, which includes all parameters in the model. In the second stage, we learn $\vtheta$ using MLE~\cite{kulesza2012determinantal}
\begin{equation}
\label{dpp_obj}
\vtheta^* = \argmax_{\vtheta} \sum_{i}\log\{P(\vz^{(i)*} \subset \cZ^{(i)}; \mL^{(i)}(\vtheta))\},
\end{equation}
where $i$ indexes the target subset, ground set, and $\mL$ matrix of the $i$-th video. We optimize $\vtheta$ with stochastic gradient descent.

\subsection{Generating shot-based summaries for vsLSTM and dppLSTM}
\label{shotgeneration}
As mentioned in Section~3.1 and 3.5 of the main text, the outputs of both our proposed models are on the frame level --- \textsf{vsLSTM} predicts frame-level importance scores, while \textsf{dppLSTM} selects a subset of keyframes using approximate MAP inference~\cite{buchbinder2015tight}. To compare with the user-annotated keyshots in \textbf{SumMe} and \textbf{TVSum} for evaluation, we convert the outputs into keyshot-based summaries.

For \textsf{vsLSTM}, we directly apply the conversion in Section~\ref{score}. We set the threshold of the total duration of keyshots to be 15\% of the original video length (for both datasets), following the protocols in~\cite{Song2015TVSum,gygli2014creating,Gygli2015video}.

For \textsf{dppLSTM}, we apply the conversion in Section~\ref{keyframe}. In practice, DPP inference usually leads to \emph{high precision} yet \emph{low recall}; i.e., the resulting total duration of keyshots may be far below the threshold (on average, 10\%). We thus add in few more keyshots by utilizing the scalar output of the MLP $f_I(\cdot)$, following the procedure in Section~\ref{score}. The MLP $f_I(\cdot)$ is pre-trained using the frame-level importance scores (cf. Section~3.4 of the main text) and conveys a certain notion of importance even after fine-tuning with the DPP objective.

\section{Comparing different network structures for~\textsf{dppLSTM}}
\label{sStruct}
The network structure of~\textsf{dppLSTM} (cf. Fig. 3 of the main text) involves two MLPs --- the MLP $f_I(\cdot)$ outputting $y_t$ for frame-level importance and the MLP $f_S(\cdot)$ outputting $\vphi_t$ for similarity.

\begin{figure}[t]
\centering
\includegraphics[height=5.5cm]{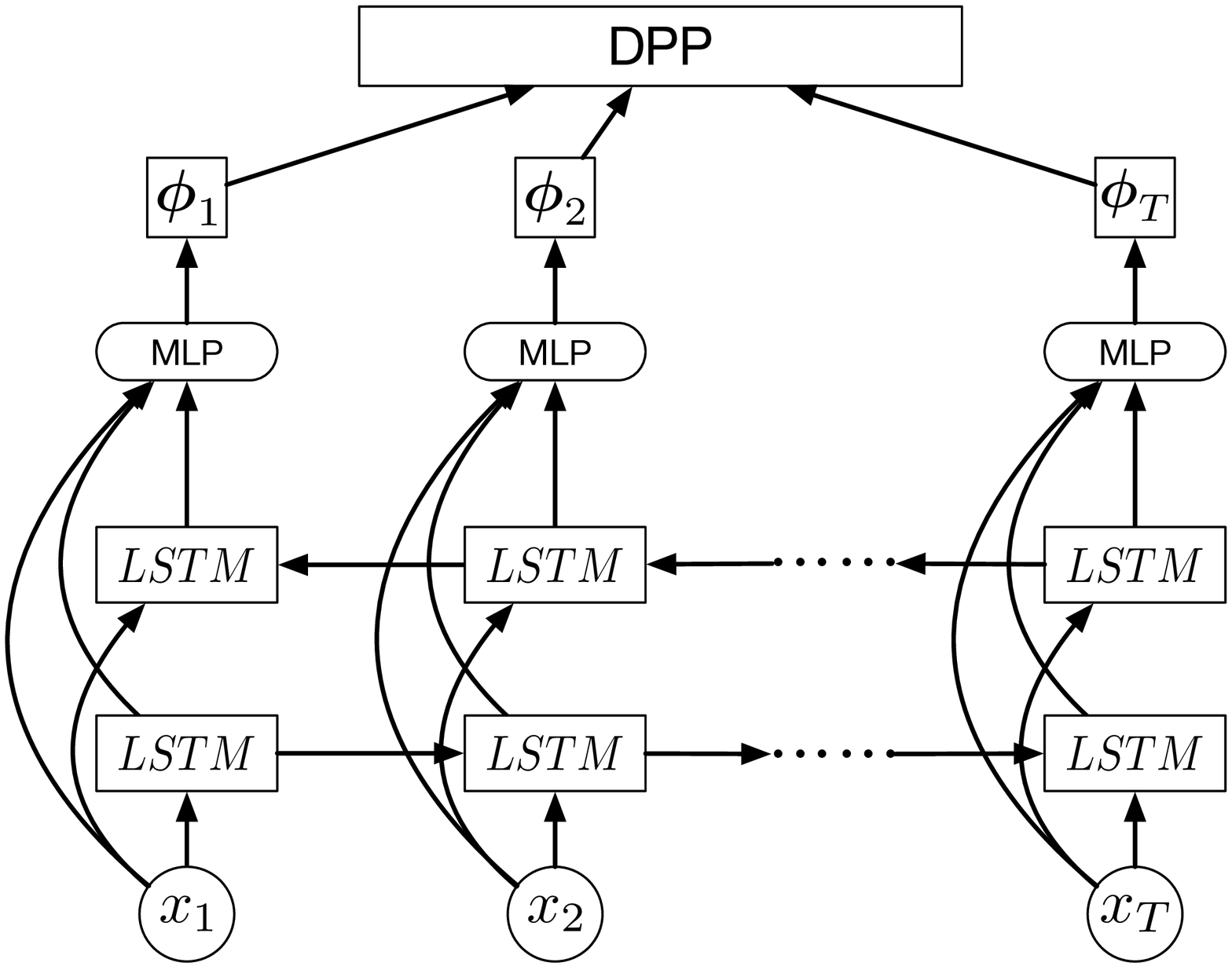}
\caption{\small Our \textsf{dppLSTM-single} model. It is similar to \textsf{dppLSTM} (Fig.~3 in the main text) but learns only a single MLP $f_S(\cdot)$ and then stacks with a DPP.}
\label{fig:dppLSTM-single}
\end{figure}

In this section, we compare with another LSTM-based model that learns only a single MLP $f_S(\cdot)$ and then stacks with a DPP. We term this model as ~\textsf{dppLSTM-single}. See Fig.~\ref{fig:dppLSTM-single} for illustration. \textsf{dppLSTM-single} also outputs a set of keyframes and is likely to generate a keyshot-based summary of an insufficient duration (similar to \textsf{dppLSTM} in Section~\ref{shotgeneration}). We thus add in few more keyshots by utilizing the diagonal values of $\mL$ as frame-level scores, following~\cite{zhang2016summary}.

Table~\ref{dppLSTMstructure} compares the performance of the two network structures, and \textsf{dppLSTM} obviously outperforms \textsf{dppLSTM-single}. As a well-learned DPP model should capture the notions of both quality (importance) and diversity~\cite{kulesza2012determinantal}, we surmise that 
separately modeling the two factors would benefit, especially when the model of each factor can be pre-trained (e.g, the MLP $f_I(\cdot)$ in \textsf{dppLSTM}).

\begin{table}[t]
\centering
\small
\caption{Comparison between \textsf{dppLSTM} and \textsf{dppLSTM-single} on different settings.}
\label{dppLSTMstructure}
\begin{tabular}{c|l|c|c|c}
\hline 
Dataset & Method & \textsf{Canonical}& \textsf{Augmented} & \textsf{Transfer} \\
\hline 
\multirow{2}{*}{\textbf{SumMe}} 
& \textsf{dppLSTM} & 38.6\scriptsize{$\pm$0.8}& \color{red}{{42.9}}\scriptsize{$\pm$0.5}& 41.8\scriptsize{$\pm$0.5}  \\ \cline{2-5}
& \textsf{dppLSTM-single} & 37.5\scriptsize{$\pm$0.9} & 41.4\scriptsize{$\pm$0.8} & 40.3\scriptsize{$\pm$0.9}  \\ \cline{2-5}

\hline \hline
\multirow{2}{*}{\textbf{TVSum}} 
& \textsf{dppLSTM} & {54.7\scriptsize{$\pm$0.7}}& \color{red}{{59.6}}\scriptsize{$\pm$0.4}& {58.7\scriptsize{$\pm$0.4}}  \\ \cline{2-5}
& \textsf{dppLSTM-single} &  53.9\scriptsize{$\pm$0.9} & 57.5\scriptsize{$\pm$0.7} & 56.2\scriptsize{$\pm$0.8}   \\ \cline{2-5}
\hline
\end{tabular}
\end{table}

\section{Other implementation details}
In this section, we provide the implementation details for both the proposed models (\textsf{vsLSTM}, \textsf{dppLSTM}) and baselines (\textsf{MLP-Frame}, \textsf{MLP-Shot}).

\label{sOther}
\subsection{Input signal}
For \textsf{vsLSTM}, \textsf{dppLSTM}, and \textsf{MLP-Frame}, which all take frame features as inputs, we uniformly subsample the videos to 2 fps\footnote{For videos with slow varying
contents such as SumMe/TVSum, this scheme seems adequate. For OVP and YouTube, even 1 fps is sufficient for fairly good summarization~\cite{de2011vsumm,gong14diverse,zhang2016summary}.}. The concatenated feature (of a 5-frame window) to \textsf{MLP-Frame} is thus equivalent to taking a 2-second span into consideration. For \textsf{MLP-Shot}, we perform KTS~\cite{potapov2014category} to segment the video into shots (disjoint intervals), where each shot is around 5 seconds on average. 

\subsection{Network structures}

$f_I(\cdot)$ and $f_S(\cdot)$ are implemented by one-hidden-layer MLPs, while \textsf{MLP-Shot} and \textsf{MLP-Frame} are two-hidden-layer MLPs. For all models, we set the size of each hidden layer of MLPs, the number of hidden units of each unidirectional LSTM, and the output dimension of the MLP $f_S(\cdot)$ all to be 256. 
We apply the sigmoid activation function to all the hidden units as well as the output layer of \textsf{MLP-Shot}, \textsf{MLP-Frame}, and $f_I(\cdot)$. The output layer of $f_S(\cdot)$ are of linear units. 
We run for each setting and each testing fold (cf. Section~4.2 of the main text) 5 times and report the average and standard deviation.

\subsection{Learning objectives}
For \textsf{MLP-Frame}, \textsf{MLP-Shot}, \textsf{vsLSTM}, and the first stage of \textsf{dppLSTM}, we use the square loss. For \textsf{dppLSTM-single} and the second stage of \textsf{dppLSTM}, we use the likelihood (cf. (\ref{dpp_obj})). 

\subsection{Stopping criteria}

For all our models, we stop training after $K$ consecutive epochs with descending summarization F-score on the validation set. We set $K = 5$.

\section{Additional discussions on video summarization}
\label{Disc_VS}

Video summarization is essentially a structured prediction problem and heavily relies on how to model/capture the sequential (or temporal) structures underlying videos. In this work, we focus on modeling the structures making sequentially inter-dependent decisions at three levels: (a) realizing boundaries of sub-events/shots; (b) removing redundant nearby shots/frames; (c) retaining temporally distant events despite being visually similar (cf. the motivating example of ``leave home'' in Section 1 of the main text). Essentially, any decision including or excluding frames is dependent on other decisions made on a temporal line.

\end{document}